\documentclass[hidelinks]{article}

\usepackage[preprint]{neurips_2023}
\usepackage{booktabs, graphicx, subcaption, pgfplots}
\usepackage[inline]{enumitem}
\usepackage[utf8]{inputenc} % allow utf-8 input
\usepackage[T1]{fontenc}    % use 8-bit T1 fonts
\usepackage{hyperref}       % hyperlinks
\usepackage{url}            % simple URL typesetting
\usepackage{booktabs}       % professional-quality tables
\usepackage{amsfonts}       % blackboard math symbols
\usepackage{nicefrac}       % compact symbols for 1/2, etc.
\usepackage{microtype}      % microtypography
\usepackage{xcolor}         % colors

\usepackage{natbib}

\title{Can a student Large Language Model perform as well as it's teacher?}

\author{
    Sia Gholami \\
    The Institute of Electrical and Electronics Engineers, Member IEEE \\
    \texttt{gholami@ieee.org} \\
    \And 
    Marwan Omar \\
    Illinois Institute of Technology \\
    \texttt{momar3@iit.edu} \\
}

\begin{document}

\maketitle

\begin{abstract}
The burgeoning complexity of contemporary deep learning models, while achieving unparalleled accuracy, has inadvertently introduced deployment challenges in resource-constrained environments. Knowledge distillation, a technique aiming to transfer knowledge from a high-capacity "teacher" model to a streamlined "student" model, emerges as a promising solution to this dilemma. This paper provides a comprehensive overview of the knowledge distillation paradigm, emphasizing its foundational principles such as the utility of soft labels and the significance of temperature scaling. Through meticulous examination, we elucidate the critical determinants of successful distillation, including the architecture of the student model, the caliber of the teacher, and the delicate balance of hyperparameters. While acknowledging its profound advantages, we also delve into the complexities and challenges inherent in the process. Our exploration underscores knowledge distillation's potential as a pivotal technique in optimizing the trade-off between model performance and deployment efficiency.
\end{abstract}

\section{Introduction}

In recent years, the landscape of deep learning has been characterized by models that are increasingly large and intricate. While such models, often boasting billions of parameters, consistently set new benchmarks in accuracy, their computational intensity presents deployment challenges, especially in environments with limited computational resources, such as edge devices~\citep{tao2020challenges}. Knowledge distillation offers a viable solution to this quandary, facilitating the transfer of knowledge from a sophisticated, high-capacity "teacher" model to a more compact "student" model, aiming to retain as much of the performance as possible~\citep{hinton2015distilling}.

Central to knowledge distillation is the principle that learning can be enhanced when models are trained not just on hard labels but also on the richer, probabilistic outputs of a teacher model. These soft labels can be perceived as capturing the teacher's confidence distribution across classes, providing nuanced insights which hard labels might overlook~\citep{buciluǎ2006model}.

A critical component of this approach is temperature scaling, which modulates the granularity of these soft labels. The temperature parameter, introduced by~\cite{hinton2015distilling}, plays a pivotal role in controlling the "sharpness" of the teacher's output distributions, thus influencing the quality of the information relayed to the student model.

The training of the student model is then typically guided by a weighted loss function that balances between the conventional cross-entropy loss and the divergence from the teacher's outputs, usually measured using Kullback-Leibler divergence~\citep{lopez2015unifying}.

However, the process is not without complexities. The optimal architecture of the student model, the quality of the teacher, and the precise balance of hyperparameters are all determining factors in the success of the distillation~\citep{polino2018model}. The intricacies of these factors and their interplay remain a focal point of contemporary research.

In conclusion, knowledge distillation emerges as a key technique in the deep learning toolkit, bridging the divide between cutting-edge performance and practical, efficient deployment. Its continued exploration holds the promise of further refining and expanding its applicability across diverse domains.

To use knowledge distillation for creating efficient transformers, the process typically involves the following steps:

\begin{enumerate}
    \item Train a large, complex transformer model as the teacher model on the task of interest.
    \item Generate a dataset of examples for the task, and use the teacher model to generate predictions for each example.
    \item Train a smaller, simpler transformer model as the student model on the same task, using the predictions of the teacher model as targets.
    \item Use a combination of the original task loss and a distillation loss to train the student model. The distillation loss encourages the student model to mimic the predictions of the teacher model, rather than just trying to optimize the original task loss.
\end{enumerate}
By using knowledge distillation in this way, it is possible to create efficient transformer models that are smaller and faster than the original model, while still achieving comparable or even better performance on the task of interest.

There are several benefits to using knowledge distillation in building efficient transformers:

% \begin{itemize}
%     \item Smaller model size: Knowledge distillation can help reduce the size of transformer models, making them more efficient to deploy on resource-constrained devices such as smartphones, IoT devices, or embedded systems.

%     \item Lower computational cost: Smaller models require less computation to perform the same task, allowing for faster inference times and reduced energy consumption.

%     \item Improved accuracy: When a large and complex transformer model is used as the teacher, the student model can benefit from the teacher's knowledge and generalization ability, leading to better accuracy than training the smaller model from scratch.

%     \item Improved generalization: Since the teacher model has learned to generalize well over a large amount of data, the student model is also able to generalize well even when trained on limited data.

%     \item Transfer learning: The teacher model can be pre-trained on a large dataset and used to transfer knowledge to the student model, enabling the student model to learn from a much larger set of data than what would be feasible to train the student model from scratch.
% \end{itemize}

\begin{enumerate}
    \item Improved efficiency: Knowledge distillation allows you to create smaller, more efficient Transformer models that require fewer computational resources for training and inference. This enables faster processing and reduced memory usage, making it easier to deploy the models on resource-constrained devices like mobile phones or edge devices.

    \item Reduced energy consumption: Smaller models produced through knowledge distillation consume less energy during inference, which is crucial for battery-powered devices and sustainable AI solutions.

    \item Faster inference: The reduced size and complexity of distilled models lead to faster inference times, which is essential in situations where real-time or low-latency processing is required, such as natural language understanding in voice assistants or real-time text translation.

    \item Enhanced generalization: Knowledge distillation transfers knowledge from a large, high-performance model to a smaller model by training on soft targets. These soft targets contain richer information about the relationships between different classes, which can help the student model learn better representations and generalize better to unseen data~\citep{komodakis2017paying}.

    \item Retained performance: Despite the reduction in size and complexity, distilled Transformer models can retain much of the performance of the larger teacher models. This means you can still achieve strong results on NLP tasks while benefiting from the efficiency improvements.

    \item Cost-effective deployment: The reduced computational requirements of distilled models can lead to lower costs when deploying AI solutions, especially in cloud-based services where computation costs are determined by the resources consumed.

    \item Easier distribution and updates: Smaller models are easier to distribute and update, reducing the time and bandwidth required for users to download and install updates, which is particularly beneficial for applications on mobile devices or in areas with limited internet connectivity.
\end{enumerate}

Overall, knowledge distillation provides a powerful technique for building efficient transformers that can achieve high accuracy, generalize well, and be deployed on resource-constrained devices.

\section{Related Works}
Natural Language Processing (NLP) has been a major area of research in Artificial Intelligence and Machine Learning
since the early days of computer science~\citep{voorhees1999trec, moldovan2000structure, brill2002analysis, ferrucci2010building, gholami2021zero, gholami2022you, gholami2022create, gholami2022alexa, gholami2022flight, brand2022text, gholami2023generative}. There are several examples of using knowledge distillation to create efficient Transformer models in the literature. Here are a few notable cases:

DistilBERT~\citep{sanh2019distilbert} is a popular example of applying knowledge distillation to create a smaller version of BERT, a large-scale pre-trained Transformer model for NLP tasks. DistilBERT has 40\% fewer parameters than the original BERT model but retains 95\% of its performance on various benchmark tasks.

Sun et al.~\citep{sun2020mobilebert} introduced MobileBERT, an efficient Transformer model created using a combination of knowledge distillation and architecture search. MobileBERT is designed for on-device NLP tasks and achieves 4.3x faster inference speed and 2.4x smaller model size compared to BERT-base, while maintaining similar performance levels.

Jiao et al.~\citep{jiao2019tinybert} presented TinyBERT, another example of applying knowledge distillation to create a smaller and faster version of BERT. TinyBERT involves a two-step knowledge distillation process: general distillation on a large-scale dataset and task-specific distillation on a target task dataset. This approach results in a model that is 7.5x smaller and 9.4x faster than BERT-base while maintaining competitive performance.

Touvron et al.~\citep{touvron2021training} proposed Data-efficient Image Transformers (DeiT), which is a Transformer model for image classification. Although DeiT focuses on the vision domain, the authors used knowledge distillation from a convolutional neural network (CNN) teacher model to improve the performance of the Transformer student model. This demonstrates the potential of cross-modal knowledge distillation in creating efficient Transformer models.

Wang et al.~\citep{wang2020minilm} proposed MiniLM, a distilled version of the pre-trained BERT model. MiniLM uses a combination of self-attention knowledge distillation and intermediate-layer knowledge distillation, aiming to preserve the original model's linguistic knowledge and structural information. MiniLM achieves a substantial reduction in model size and faster inference while maintaining competitive performance on various NLP benchmarks.

Fan et al.~\citep{fan2019reducing} proposed LayerDrop, a regularization technique for efficiently training and distilling deep Transformer models. LayerDrop trains a single model that can be efficiently pruned at inference time by dropping layers, resulting in a family of smaller models with varying trade-offs between performance and efficiency. This approach can be combined with knowledge distillation to create even more efficient Transformer models.

Gao et al.~\citep{chakraborty2021recovery} introduced RocketQAv2, a distilled version of T5 (Text-to-Text Transfer Transformer) for open-domain question-answering tasks. RocketQAv2 is based on the T5 model but utilizes knowledge distillation techniques to create a smaller model that is more efficient at serving real-world applications.

These examples highlight the effectiveness of knowledge distillation in creating smaller, faster, and more efficient Transformer models while maintaining competitive performance on various NLP and vision tasks.

\section{Approach}
Knowledge distillation is a technique used in machine learning to transfer knowledge from a larger, more complex model (called the teacher model) to a smaller, simpler model (called the student model). The goal is to create a lightweight, computationally efficient student model that retains as much of the teacher model's performance as possible. This is especially useful for deploying machine learning models on resource-constrained devices or in situations where inference time and energy consumption are critical.

The knowledge distillation approach entails training a smaller, more focused model to replicate the results of a bigger, more broad language model, like GPT-3~\citep{brown2020language}. The bigger model's high-level symbolic knowledge should be reduced into a smaller, more effective model that can accurately carry out specific tasks. This method involves training a student Transformer model using knowledge distillation from a larger teacher model. The teacher model provides soft labels for the training data, which are used to train the student model. This allows the student model to learn from the teacher model's rich representation while being more efficient due to its smaller size~\citep{freitag2017ensemble}. In our experiments we used the model introduced by~\cite{gholami2023generative} (GPT-Efficio) as the teacher.

Here we study a specific approach called The figurative distiller (FD) process that involves three main components: a teacher model, a student model, and a set of rules. The teacher model is the larger, more general language model, while the student model is the smaller, more specialized model being trained. The rules define the constraints and relationships between different concepts in the modeled domain. There are three steps in the FD procedure. The student model is first trained using a collection of training examples produced by the instructor model. In order to enhance the student model's performance, a task-specific dataset is used to fine-tune it. The output of the student model is then further refined using symbolic rules to ensure it adheres to the limitations and relationships specified by the rules.

Both supervised, and unsupervised learning can be included in the FD process. The teacher model creates a collection of labeled examples in the supervised environment, subsequently utilized for training the student model. In an unsupervised scenario, the student model is trained using methods like contrastive learning utilizing a collection of unlabeled examples that the teacher model generated.

The objective of figurative distiller  is to learn a smaller, more specialized model, $f_S$, from a dataset of input/output pairs, $D = (x_1, y_1),..., (x_n, y_n)$, where $x_i$ is an input sequence, and $y_i$ is a symbolic output sequence. A pre-trained general language model, $f_G$, can carry out the same task as $f_G$ but with fewer parameters. We suggest a figurative distiller  technique to accomplish this aim, which entails training the smaller model, $f_S$, using a mix of supervised and unsupervised learning. For the figurative distiller , the loss function is as follows:

The loss function for knowledge distillation, consisting of a combination of the cross-entropy loss with the true labels and the KL-divergence loss between the teacher and student outputs, is typically written as:

\begin{equation}
    L = \alpha \times CE(y, Student(x)) + (1 - \alpha) \times T^2 \times KL( Student(\frac{x}{T}) || Teacher(\frac{x}{T}))    
\end{equation}

Where $L$ is the total loss function, $\alpha$ is a weighting factor determining the balance between the original loss and the distillation loss, $CE$ is the cross-entropy loss function, $y$ is the true labels, $Student(x)$ is the student model's predictions, $T$ is the temperature parameter used to soften probabilities, $KL$ is the Kullback-Leibler divergence, and $Teacher(x)$ is the teacher model's predictions. 

The KL-divergence and cross-entropy are both measured across all the classes, and we're summing these measurements to get a scalar loss.

The Kullback-Leibler (KL) divergence is a measure of how one probability distribution diverges from a second, expected probability distribution. For discrete probability distributions P and Q, the KL divergence is defined as:

\begin{equation}
    KL(P||Q) = \sum P(i) \times \log(\frac {P(i)}{Q(i)})    
\end{equation}

where the sum is over all possible events $i$, $P(i)$ is the probability of event $i$ under distribution $P$, and $Q(i)$ is the probability of event $i$ under distribution $Q$.

For continuous distributions, the sum is replaced by an integral over all possible outcomes.

\section{Experiments}

In this section we present the results of each of our approaches in the context of language modeling (i.e. completion tasks) and question answering.

\subsection{Results}

This section investigates the knowledge distillation techniques to compress large models into smaller ones while retaining a good deal of the performance of the original model. Some of the impacts of using knowledge distillation on a transformer model are:

\begin{enumerate}
    \item Improved Efficiency: A smaller distilled model has fewer parameters and thus requires less computational resources for inference. This makes it possible to deploy the model on devices with limited resources, such as mobile devices.
    \item Speed: The smaller model should also be faster, both in terms of training and inference times, compared to the original larger model. 
    \item Performance: Generally, the distilled model will perform worse than the original larger model, but better than a similarly-sized model trained from scratch~\citep{du2017learning}. The goal is to retain as much performance as possible given the constraints on model size.
    \item Robustness: In some cases, knowledge distillation may also increase the robustness of the model and its ability to generalize, as it learns to mimic the teacher's predictions over a wide variety of samples and not just the ground truth.
\end{enumerate}

However, it's important to note that these benefits depend on the specifics of the task, the architecture of the teacher and student models, and the training procedure used. Poorly configured distillation can result in a model that performs no better or even worse than a model of the same size trained from scratch.

\begin{table}[!htbp]
\centering
\small
\caption{Performance of knowledge distillation approach on completion tasks}\label{kd-lm}
\begin{tabular}{p{0.2\linewidth} p{0.1\linewidth} p{0.15\linewidth} p{0.15\linewidth} p{0.12\linewidth} p{0.12\linewidth}}
\toprule
\textbf{Model} & \textbf{$n_{params}$} & \textbf{LAMBADA (acc)} & \textbf{LAMBADA (ppl)} & \textbf{StoryCloze (acc)} & \textbf{HellaSwag (acc)} \\ 
\midrule
GPT-3 Zero-Shot & 175B & 76.2 & 3.00 & 83.2 & 78.9   \\ 
GPT-3 One-Shot & 175B & 72.5 & 3.35 & 84.7 & 78.1   \\ 
GPT-3 Few-Shot & 175B & 86.4 & 1.92 & 87.7 & 79.3   \\ 
GPT-Efficio (teacher) & 950M & 67.1 & 9.2 & 80.5 & 72.6 \\
GPT-Efficio (student) & 320M & 52.47 & 13.53 & 61.28 & 63.52 \\
\bottomrule
\end{tabular}
\end{table}

Table~\ref{kd-lm} demonstrates the GPT-Efficio teacher and student models performance in comparison with GPT-3. 

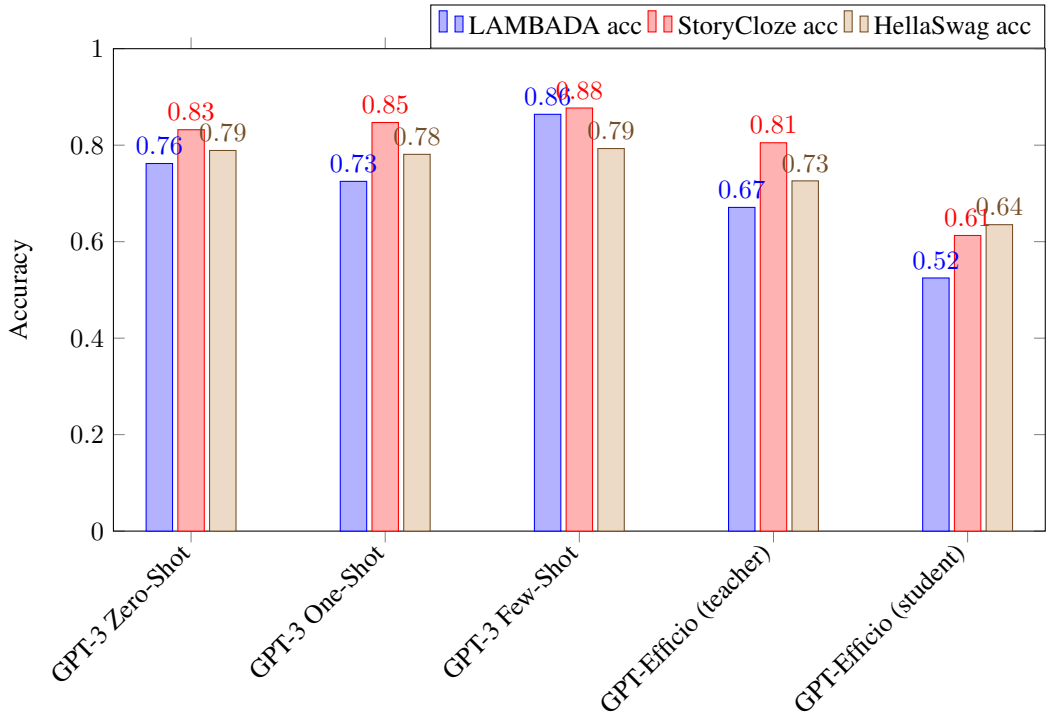
\begin{figure}[htbp]
\centering
\begin{tikzpicture}
    \begin{axis}[
        width=\linewidth,
        height=8cm,
        ybar, ymin=0, ymax=1,
        legend style={at={(0.5,-0.15)},
        anchor=north,legend columns=-1},
        ylabel={Accuracy},
        symbolic x coords={GPT-3 Zero-Shot, GPT-3 One-Shot, GPT-3 Few-Shot, GPT-Efficio (teacher), GPT-Efficio (student)},
        xtick=data,
        nodes near coords,
        x tick label style={rotate=45,anchor=east},
        legend style={at={(1,1)},
        anchor=south east,legend columns=-1}
        ]
        \addplot coordinates {(GPT-3 Zero-Shot,0.762) (GPT-3 One-Shot,0.725) (GPT-3 Few-Shot,0.864) (GPT-Efficio (teacher),0.671) (GPT-Efficio (student),0.5247)};
        \addplot coordinates {(GPT-3 Zero-Shot,0.832) (GPT-3 One-Shot,0.847) (GPT-3 Few-Shot,0.877) (GPT-Efficio (teacher),0.805) (GPT-Efficio (student),0.6128)};
        \addplot coordinates {(GPT-3 Zero-Shot,0.789) (GPT-3 One-Shot,0.781) (GPT-3 Few-Shot,0.793) (GPT-Efficio (teacher),0.726) (GPT-Efficio (student),0.6352)};
        \legend{LAMBADA acc,StoryCloze acc,HellaSwag acc}
    \end{axis}
\end{tikzpicture}
\caption{Performance of knowledge distillation approach on completion tasks}
\label{kd-lm}
\end{figure}

\begin{table}[!h]
\centering
\small
\caption{Performance of knowledge distillation approach on QA tasks}\label{kd-qa}
\begin{tabular}{p{0.2\linewidth} p{0.1\linewidth} p{0.15\linewidth} p{0.15\linewidth} p{0.12\linewidth}}
\toprule
\textbf{Model} & \textbf{$n_{params}$} & \textbf{NQ} & \textbf{WebQ} & \textbf{TriviaQA}\\ 
\midrule
GPT-3 Zero-Shot & 175B & 14.6 & 14.4 & 64.3   \\ 
GPT-3 One-Shot & 175B & 23.0 & 25.3 & 68.0   \\ 
GPT-3 Few-Shot & 175B & 29.9 & 41.5 & 71.2   \\ 
GPT-Efficio (teacher) & 950M & 27.5 & 40.6 & 69.2 \\
%GPT-Efficio (student) & 120M & 20.61 & 32.52 & 53.61 \\
GPT-Efficio (student) & 320M & 19.61 & 30.52 & 53.61 \\
\bottomrule
\end{tabular}
\end{table}

Table~\ref{kd-qa} shows the GPT-Efficio teacher and student models performance in comparison with GPT-3. 

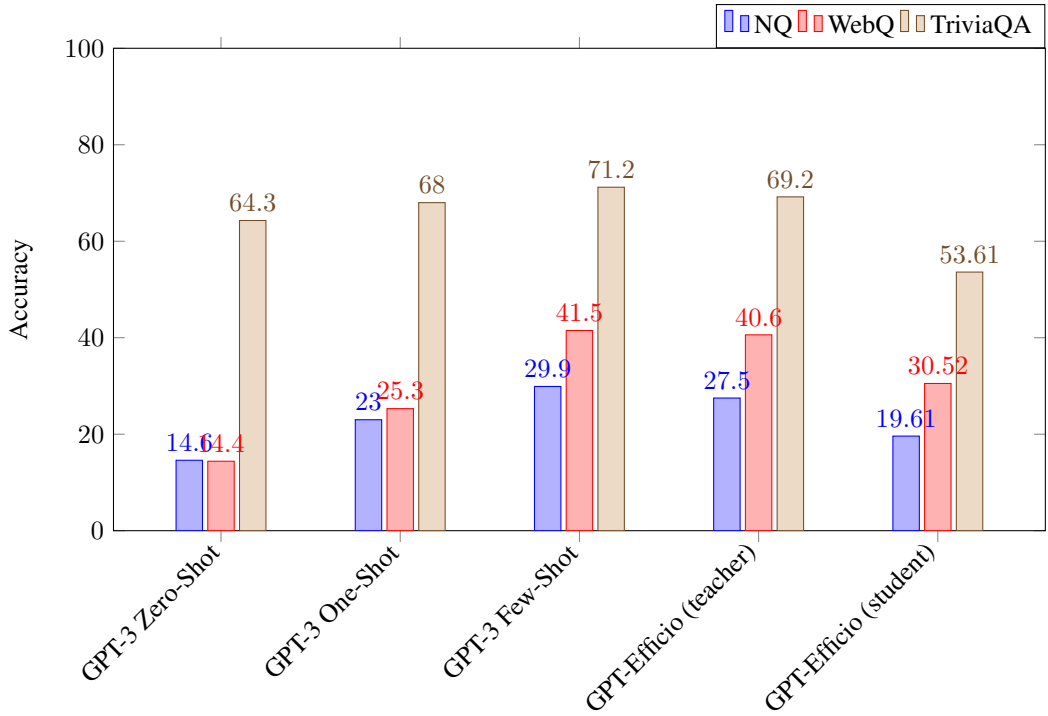
\begin{figure}
\centering
\begin{tikzpicture}
\begin{axis}[
    width=\linewidth,
    height=8cm,
    ybar, ymin=0, ymax=100,
    enlarge x limits=0.15,
    legend style={at={(0.5,-0.15)},
      anchor=north,legend columns=-1},
    ylabel={Accuracy},
    symbolic x coords={GPT-3 Zero-Shot, GPT-3 One-Shot, GPT-3 Few-Shot, GPT-Efficio (teacher), GPT-Efficio (student)},
    xtick=data,
    nodes near coords,
    nodes near coords align={vertical},
    x tick label style={rotate=45,anchor=east},
    legend style={at={(1,1)},
        anchor=south east,legend columns=-1}
]
\addplot coordinates {(GPT-3 Zero-Shot,14.6) (GPT-3 One-Shot,23.0) (GPT-3 Few-Shot,29.9) (GPT-Efficio (teacher),27.5) (GPT-Efficio (student),19.61)};
\addplot coordinates {(GPT-3 Zero-Shot,14.4) (GPT-3 One-Shot,25.3) (GPT-3 Few-Shot,41.5) (GPT-Efficio (teacher),40.6) (GPT-Efficio (student),30.52)};
\addplot coordinates {(GPT-3 Zero-Shot,64.3) (GPT-3 One-Shot,68.0) (GPT-3 Few-Shot,71.2) (GPT-Efficio (teacher),69.2) (GPT-Efficio (student),53.61)};
\legend{NQ, WebQ, TriviaQA}
\end{axis}
\end{tikzpicture}
\caption{Performance of knowledge distillation approach on QA tasks.}
\label{fig:kd-qa}
\end{figure}

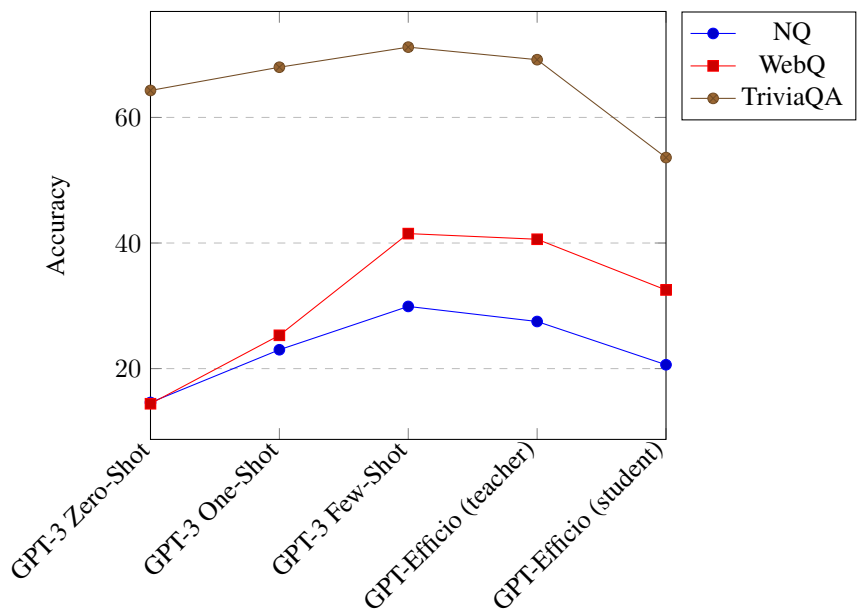
\begin{figure}
\centering
\begin{tikzpicture}
\begin{axis}[
    title= {Performance of knowledge distillation approach on QA tasks},
    ylabel={Accuracy},
    xmin=1, xmax=5,
    xtick=data,
    xticklabels={GPT-3 Zero-Shot, GPT-3 One-Shot, GPT-3 Few-Shot, GPT-Efficio (teacher), GPT-Efficio (student)},
    x tick label style={rotate=45,anchor=east},
    legend pos=outer north east,
    ymajorgrids=true,
    grid style=dashed,
]
\addplot+[sharp plot]
coordinates {(1,14.6) (2,23.0) (3,29.9) (4,27.5) (5,20.61)};
\addplot+[sharp plot]
coordinates {(1,14.4) (2,25.3) (3,41.5) (4,40.6) (5,32.52)};
\addplot+[sharp plot]
coordinates {(1,64.3) (2,68.0) (3,71.2) (4,69.2) (5,53.61)};
\legend{NQ, WebQ, TriviaQA}
\end{axis}
\end{tikzpicture}
\caption{Performance of knowledge distillation approach on QA tasks.}
\label{fig:kd-qa-line}
\end{figure}

%\clearpage

\section{Analysis}
In the context of knowledge distillation, there are several key hyperparameters that can impact the performance of the student model:

\begin{enumerate}
    \item Temperature ($T$): The temperature is a parameter in the softmax function used to "soften" the outputs of the teacher and student models during the distillation process. A higher temperature results in a softer probability distribution over classes, making the distillation process more effective by highlighting the relationships between different classes.~\citep{hinton2015distilling}. However, setting the temperature too high could result in over-softening and lead to a loss of valuable information.
    
    \item Distillation Coefficient ($\alpha$): This is the weight given to the original loss (typically cross-entropy loss with the true labels) in the combined loss function. The balance between this original loss and the distillation loss (KL-divergence between the teacher and student outputs) is crucial. If $\alpha$ is set too high, the student may focus too much on matching the true labels and not enough on learning from the teacher's predictions.~\citep{zagoruyko2016paying}.
    
    \item Model Architecture: While not technically a hyperparameter of the distillation process itself, the architecture of the student model can significantly impact its performance. If the student model is too small, it may not have enough capacity to learn from the teacher effectively. Conversely, if it's too large, the benefits of distillation (such as increased efficiency and speed) may not be realized.

    \item Learning Rate and Other Training Parameters: As with any machine learning model, the learning rate and other training parameters can significantly impact the performance of the student model.
    
    \item Number of Distillation Epochs: The number of epochs the student is trained to match the teacher's outputs can also affect performance. Too few epochs might result in underfitting, while too many might lead to overfitting.

\end{enumerate}

As always, these hyperparameters may need to be tuned depending on the specifics of the task, the dataset, and the architectures of the teacher and student models.

In this section we investigate the effects of Temperature ($T$) on the student model. The temperature parameter $T$ in knowledge distillation plays a key role in controlling the "sharpness" of the probability distribution output by the teacher model.

In the context of language modeling, if $T$ is low (closer to 1 or less), the teacher model's output probabilities will be more "sharp" or "peaky", meaning that the model will assign high probabilities to a few select words and very low probabilities to the rest. This can make it difficult for the student model to learn more nuanced behaviors from the teacher model because the gradient of the loss function becomes sparse and learning slows down.

On the other hand, if $T$ is high, the teacher model's output probabilities become more uniform or "soft". This means the model assigns more evenly distributed probabilities across a larger set of words. This can be beneficial in cases with multiple correct answers, as it encourages the student model to consider a wider range of possibilities rather than fixating on a single correct answer. It essentially provides a richer information set to the student model during training.

However, setting $T$ too high can also have downsides. If the teacher's output probabilities become too uniform, the student model might struggle to identify the most likely and less likely words. This could lead to underfitting, where the student model becomes less accurate because it's less confident in its predictions.

So, the temperature $T$ should be set in such a way that it balances the need for the student model to learn nuanced behaviors from the teacher model while also ensuring the student model can discern between more and less likely predictions. Fine-tuning this parameter might require some experimentation or validation on a separate dev set.

\begin{table}[!htbp]
\centering
\small
\caption{Analysis of the effects of hyperparameter $T$ on completion tasks}\label{kd-anal-lm}
\begin{tabular}{p{0.2\linewidth} p{0.05\linewidth} p{0.1\linewidth} p{0.15\linewidth} p{0.15\linewidth} p{0.12\linewidth} p{0.12\linewidth}}
\toprule
\textbf{Model} & \textbf{$T$} & \textbf{$n_{params}$} & \textbf{LAMBADA (acc)} & \textbf{LAMBADA (ppl)} & \textbf{StoryCloze (acc)} & \textbf{HellaSwag (acc)} \\ 
\midrule
GPT-3 Zero-Shot & - &  175B & 76.2 & 3.00 & 83.2 & 78.9   \\ 
GPT-3 One-Shot & - & 175B & 72.5 & 3.35 & 84.7 & 78.1   \\ 
GPT-3 Few-Shot & - & 175B & 86.4 & 1.92 & 87.7 & 79.3   \\ 
GPT-Efficio (teacher) & - & 950M & 67.1 & 9.2 & 80.5 & 72.6 \\
%GPT-Efficio (student) & 1 & 320M & 52.47 & 18.53 & 71.28 & 63.52 \\
GPT-Efficio (student) & 1 & 320M & 52.47 & 13.53 & 61.28 & 63.52 \\
%GPT-Efficio (student) & 2 & 320M & 49.40 & 22.69 & 65.29 & 57.59 \\
GPT-Efficio (student) & 2 & 320M & 49.40 & 14.69 & 59.29 & 60.59 \\
\bottomrule
\end{tabular}
\end{table}

Table~\ref{kd-anal-lm} demonstrates the GPT-Efficio teacher and student models performance with various $T$ values in comparison with GPT-3. 
\begin{figure}
\begin{tikzpicture}
\begin{axis}[
    ybar, ymin=0, ymax=100,
    bar width=.2cm,
    width=\textwidth,
    height=.5\textwidth,
    legend style={at={(0.5,-0.15)},
      anchor=north,legend columns=-1},
    ylabel={Accuracy},
    symbolic x coords={GPT-3 Zero-Shot, GPT-3 One-Shot, GPT-3 Few-Shot, GPT-Efficio T-, GPT-Efficio T1, GPT-Efficio T2},
    xtick=data,
    x tick label style={rotate=45,anchor=east},
    ymin=0,
    enlarge x limits=0.1,
    nodes near coords,
    nodes near coords align={vertical},
    every node near coord/.append style={rotate=90, anchor=west},
    legend style={at={(1,1)},
        anchor=south east,legend columns=-1}
    ]
\addplot coordinates {
    (GPT-3 Zero-Shot, 76.2)
    (GPT-3 One-Shot, 72.5)
    (GPT-3 Few-Shot, 86.4)
    (GPT-Efficio T-, 67.1)
    (GPT-Efficio T1, 52.47)
    (GPT-Efficio T2, 49.40)
};
\addlegendentry{LAMBADA (acc)}
\addplot coordinates {
    (GPT-3 Zero-Shot, 83.2)
    (GPT-3 One-Shot, 84.7)
    (GPT-3 Few-Shot, 87.7)
    (GPT-Efficio T-, 80.5)
    (GPT-Efficio T1, 61.28)
    (GPT-Efficio T2, 59.29)
};
\addlegendentry{StoryCloze (acc)}
\addplot coordinates {
    (GPT-3 Zero-Shot, 78.9)
    (GPT-3 One-Shot, 78.1)
    (GPT-3 Few-Shot, 79.3)
    (GPT-Efficio T-, 72.6)
    (GPT-Efficio T1, 63.52)
    (GPT-Efficio T2, 60.59)
};
\addlegendentry{HellaSwag (acc)}
\end{axis}
\end{tikzpicture}
\caption{Analysis of the effects of hyperparameter $T$ on completion tasks}
\end{figure}
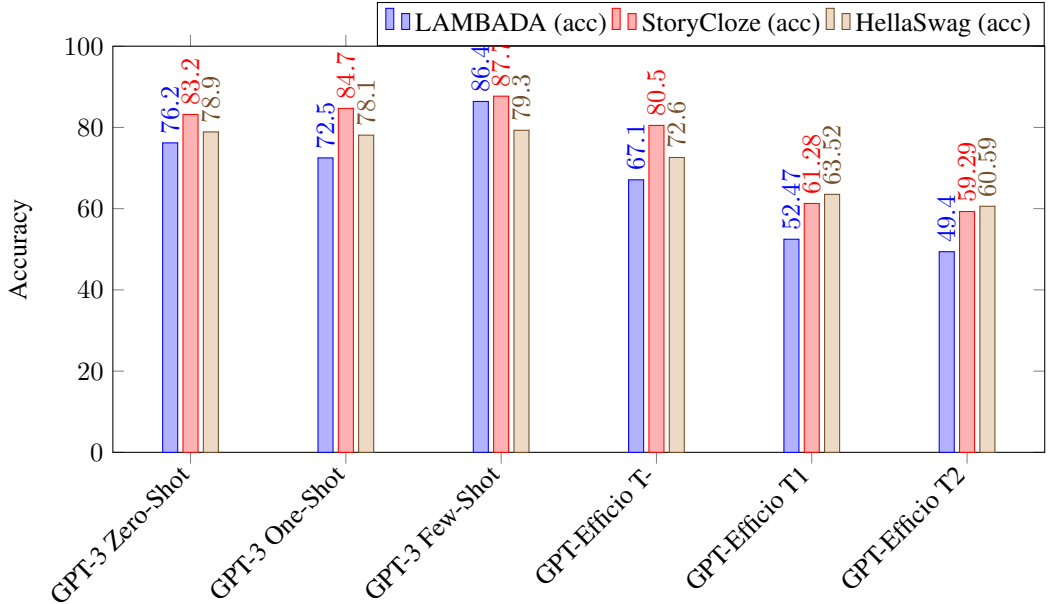

\begin{table}[!htbp]
\centering
\small
\caption{Analysis of the effects of hyperparameter $T$ on QA tasks}\label{kd-anal-qa}
\begin{tabular}{p{0.2\linewidth} p{0.05\linewidth} p{0.1\linewidth} p{0.15\linewidth} p{0.15\linewidth} p{0.12\linewidth}}
\toprule
\textbf{Model} & \textbf{$T$} & \textbf{$n_{params}$} & \textbf{NQ} & \textbf{WebQ} & \textbf{TriviaQA}\\ 
\midrule
GPT-3 Zero-Shot & - & 175B & 14.6 & 14.4 & 64.3   \\ 
GPT-3 One-Shot & - & 175B & 23.0 & 25.3 & 68.0   \\ 
GPT-3 Few-Shot & - & 175B & 29.9 & 41.5 & 71.2   \\ 
GPT-Efficio (teacher) & - & 950M & 27.5 & 40.6 & 69.2 \\
%GPT-Efficio (student) & 1 & 120M & 20.61 & 32.52 & 53.61 \\
GPT-Efficio (student) & 1 & 320M & 19.61 & 30.52 & 53.61 \\
%GPT-Efficio (student) & 2 & 120M & 18.19 & 29.87 & 44.50 \\
GPT-Efficio (student) & 2 & 320M & 17.19 & 27.87 & 48.50 \\
\bottomrule
\end{tabular}
\end{table}

Table~\ref{kd-anal-qa} shows the GPT-Efficio performance teacher and student models performance with various $T$ values in comparison with GPT-3. 

\begin{figure}
\begin{tikzpicture}
\begin{axis}[
    ybar, ymin=0, ymax=100,
    bar width=.2cm,
    width=\textwidth,
    height=.5\textwidth,
    legend style={at={(0.5,-0.15)},
      anchor=north,legend columns=-1},
    ylabel={Accuracy},
    symbolic x coords={GPT-3 Zero-Shot, GPT-3 One-Shot, GPT-3 Few-Shot, GPT-Efficio T-, GPT-Efficio T1, GPT-Efficio T2},
    xtick=data,
    x tick label style={rotate=45,anchor=east},
    ymin=0,
    enlarge x limits=0.1,
    nodes near coords,
    nodes near coords align={vertical},
    every node near coord/.append style={rotate=90, anchor=west},
    legend style={at={(1,1)},
        anchor=south east,legend columns=-1}
    ]
\addplot coordinates {
    (GPT-3 Zero-Shot, 14.6)
    (GPT-3 One-Shot, 23.0)
    (GPT-3 Few-Shot, 29.9)
    (GPT-Efficio T-, 27.5)
    (GPT-Efficio T1, 19.61)
    (GPT-Efficio T2, 17.19)
};
\addlegendentry{NQ}
\addplot coordinates {
    (GPT-3 Zero-Shot, 14.4)
    (GPT-3 One-Shot, 25.3)
    (GPT-3 Few-Shot, 41.5)
    (GPT-Efficio T-, 40.6)
    (GPT-Efficio T1, 30.52)
    (GPT-Efficio T2, 27.87)
};
\addlegendentry{WebQ}
\addplot coordinates {
    (GPT-3 Zero-Shot, 64.3)
    (GPT-3 One-Shot, 68.0)
    (GPT-3 Few-Shot, 71.2)
    (GPT-Efficio T-, 69.2)
    (GPT-Efficio T1, 53.61)
    (GPT-Efficio T2, 48.50)
};
\addlegendentry{TriviaQA}
\end{axis}
\end{tikzpicture}
\caption{Analysis of the effects of hyperparameter $T$ on QA tasks}
\end{figure}
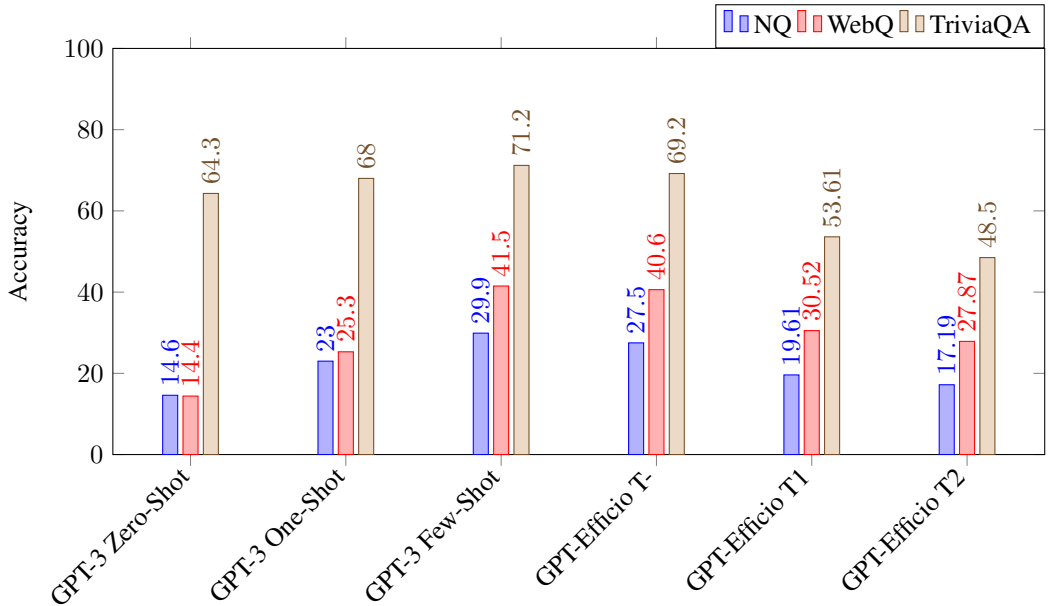

\section{Limitations}

While knowledge distillation is a powerful tool, it's not without its limitations and challenges. Here are a few to consider:

\begin{enumerate}
    \item Performance Gap: The performance of the distilled student model, although better than a similarly-sized model trained from scratch, typically doesn't reach the performance level of the larger teacher model. There is usually a trade-off between model size and accuracy.

    \item Dependence on a Good Teacher Model: The effectiveness of knowledge distillation heavily depends on the quality of the teacher model. The student model can only be as good as the teacher model allows. If the teacher model isn't well-trained or doesn't perform well, the student model is unlikely to perform well either.

    \item Hyperparameter Sensitivity: The process of knowledge distillation requires careful tuning of several hyperparameters, such as the temperature parameter and the weighting between the original loss and the distillation loss. Finding the right settings can be tricky and might require a lot of experimentation.

    \item Computational Overhead: Although the student model is smaller and more efficient, the distillation process itself requires the teacher model to generate predictions for the data, which could be computationally expensive, especially for large models and datasets.

    \item Opaque Process: The process of knowledge distillation is somewhat opaque and difficult to interpret. It's not always clear why a certain student model learns effectively from a teacher model, or why certain hyperparameters work better than others.

    \item Student Model Capacity: There's also a limit to how much a smaller student model can learn from a large teacher model. If the student model's capacity is too small, it may not effectively learn the teacher's knowledge.

    \item Overfitting Risk: If the teacher model has overfit to the training data, it's possible that the student model might learn these overfit predictions, leading to poor generalization to new data. 
\end{enumerate}

Despite these limitations, knowledge distillation can still be a very useful technique, especially when dealing with constraints on computational resources or when deploying models in real-world applications where efficiency is key. 

\section{Future Work}
There are several directions that future work on knowledge distillation could take to further improve this technique and its application in various fields:

\begin{itemize}
    \item Improved Understanding of Distillation Dynamics: Further research is needed to understand the dynamics of knowledge transfer during distillation. For example, understanding which aspects of the teacher's knowledge are most effectively transferred and why could help optimize the process.

    \item Automated Hyperparameter Tuning: Given the sensitivity of the distillation process to hyperparameters like the temperature and the weighting between the original loss and distillation loss, developing methods for automatic or more efficient hyperparameter tuning could be beneficial.

    \item Advanced Distillation Techniques: Exploring advanced distillation techniques beyond the standard approach could lead to better results. This could involve novel loss functions, training methods, or types of teacher-student relationships.

    \item Multi-Teacher Distillation: The idea of distilling knowledge from multiple teacher models into a single student model is an interesting area for exploration. This could potentially combine the strengths of various models into a single efficient student model.

    \item Domain-Specific Adaptations: Adapting and optimizing knowledge distillation techniques for specific domains or tasks could also be a valuable avenue for future work. Different tasks might benefit from different distillation strategies.

    \item Privacy and Security in Distillation: As distillation involves transferring knowledge from a teacher model, there could be concerns about privacy and security, especially when the teacher model has been trained on sensitive data. Future work could look at how to ensure that distillation does not leak sensitive information.

    \item Understanding Limitations and Failures: More research on when and why knowledge distillation fails could help in developing more robust and reliable distillation methods.
\end{itemize}

The field is evolving rapidly and the relevance of these directions could change as more research is done and newer techniques are developed.

\section{Conclusion}
The meteoric rise in the depth and complexity of neural architectures has underscored the pressing need for efficient deployment strategies in real-world scenarios. Knowledge distillation has been illuminated as a beacon in this quest, offering a method to harness the prowess of advanced models within more manageable, deployment-friendly confines. Throughout this paper, we dissected the intricate facets of this technique, from the nuances of soft label utilization to the pivotal role of temperature scaling, and highlighted the myriad determinants that influence the success of the distillation process.

However, as with many solutions in the realm of deep learning, knowledge distillation is not devoid of challenges. The precise interplay of the teacher-student dynamic, optimal hyperparameter settings, and the architecture intricacies of the student model underscore the multi-dimensional nature of the process. Further, the trade-off between model efficiency and performance, while alleviated, remains a factor to be carefully navigated.

Looking forward, it is evident that the domain of knowledge distillation offers a rich tapestry of research avenues. As we continue to push the boundaries of model performance, the concurrent pursuit of efficiency becomes indispensable. Knowledge distillation, in this landscape, stands as both a testament to our advancements and a promising frontier for future exploration. It encapsulates the essence of contemporary deep learning research: the marriage of performance and pragmatism, aimed at bringing cutting-edge AI solutions closer to real-world applicability.

\bibliographystyle{plainnat}
\bibliography{main}

\end{document}